# Graph partition strategies for generalized mean field inference

**Eric P. Xing**
Computer Science Division
University of California
Berkeley, CA 94720

**Michael I. Jordan**
Computer Science and Statistics
University of California
Berkeley, CA 94720

**Stuart Russell**
Computer Science Division
University of California
Berkeley, CA 94720

## Abstract

An autonomous variational inference algorithm for arbitrary graphical models requires the ability to optimize variational approximations over the space of model parameters as well as over the choice of tractable families used for the variational approximation. In this paper, we present a novel combination of graph partitioning algorithms with a generalized mean field (GMF) inference algorithm. This combination optimizes over disjoint clustering of variables and performs inference using those clusters. We provide a formal analysis of the relationship between the graph cut and the GMF approximation, and explore several graph partition strategies empirically. Our empirical results provide rather clear support for a weighted version of MinCut as a useful clustering algorithm for GMF inference, which is consistent with the implications from the formal analysis.

## 1 Introduction

What are the prospects for fully autonomous algorithms for variational inference in graphical models? Recent years have seen an increasingly systematic treatment of an increasingly flexible range of algorithms for variational inference. In particular, the cluster variational framework has provided a range of algorithms that extend the basic "belief propagation" framework (Yedidia et al., 2000). Similarly, general clusters of variables are also allowed in recent treatments of structured mean field algorithms (Wiegerinck, 2000). Empirical results have shown that both kinds of generalization can yield more effective algorithms.

While these developments provide much needed flexibility for the design of effective algorithms, they also raise a new question—how are the clusters to be chosen? Until now, this issue has generally been left in the hands of the algorithm designer; moreover, the designer has been provided with little beyond intuition for making these choices. For some graphical model architectures, there are only a few natural choices, and these can be explored manually. In general, however, we wish to envisage a general piece of software for variational inference which can be asked to perform inference for an arbitrary graph. In this setting, it is essential to begin to explore automatic methods for choosing clusters.

In the setting of structured mean field algorithms (Jordan et al., 1999, Ghahramani and Jordan, 1997) it is meaningful to consider *disjoint* clusters, and in Xing et al. (2003) we have proposed a *generalized mean field* (GMF) algorithm for inference based on this assumption, noting that the assumption of disjoint clusters leads to a simple set of inference equations that can be easily implemented. Disjoint clusters have another virtue as well, which is the subject of the current paper—they open the door to a role for graph partitioning algorithms in choosing clusters for inference.

There are several intuitions that support a possible role for graph partitioning algorithms in the autonomous choice of clusters for graphical model inference. The first is that minimum cuts are to be preferred, so that as much as possible of the probabilistic dependence is captured within clusters. It also seems likely that the values of parameters should matter because they often reflect the ?coupling strength? of the probabilistic dependences among random variables. Another intuition is that maximum cuts should be preferred, because in this case the mean field acting across a large cut may have an expectation that is highly concentrated, a situation which corresponds to the basic assumption underlying mean field methods. Again, specific values of parameters should matter.

In this paper we provide a preliminary formal analysis and a thoroughgoing empirical exploration of these is-sues. We present a theorem that relates the weight of the graph cut to the quality of the bound of GMF approximation, and study random graphs and a variety of settings of parameter values. We compare several different kinds of partitioning algorithms empirically. As we will show, our results turn out to provide rather clear support for a clustering algorithm based on minimal cut, which is consistent with implications drawn from the formal analysis. A more general version of the GMF algorithm that allows non-factorizable potentials is also provide in this paper and possible extensions motivated by our formal analysis are discussed.

## 2 The generalized mean field (GMF) algorithm

*Mean field approximation* refers to a class of variational approximation methods that approximate the true posterior distribution $p(\mathbf{X})$ with a simpler distribution $q(\mathbf{X})$, for which it is feasible to do exact inference. Such distributions are referred to as *tractable families*. A tractable family usually corresponds to a subgraph of a graphical model.

The generalized mean field algorithm developed in Xing et al. (2003) uses a disjoint clustering of the domain variables, $\mathcal{C} = \{C_1, C_2, \dots, C_I\}$, where $C_i$ refers to the set of indices of nodes in the $i$th cluster, to define a subgraph made up of tractable connected components of clusters of nodes. This results in a GMF approximation to the joint posterior that can be expressed as a product of (tractable) cluster marginals: $q(\mathbf{X}) = \prod q_i(\mathbf{X}_{C_i})$.

The GMF algorithm developed in Xing et al. (2003) assumes the potential functions of the cliques in the graphical models to be *cluster-factorizable*, which is not always satisfiable for general distributions, for example, in case of a distribution defined by tabular potential functions. As a prelude to our investigation of graph partitioning and GMF approximation, we briefly summarize an improved version of the GMF algorithm, which subsumes the previous version in Xing et al. (2003).

We assume that the graphical model of interest admits an exponential representation, i.e.,

$$p(\mathbf{X}|\boldsymbol{\theta}) = \frac{1}{Z_p} \exp\{\sum_{\alpha \in \mathcal{A}} \theta_\alpha \phi_\alpha(\mathbf{X}_{D_\alpha})\}, \tag{1}$$

where $\mathcal{D} = \{D_\alpha | \alpha \in \mathcal{A}\}$ denotes a set of *cliques* of the graph, indexed by a set $\mathcal{A}$, where $\boldsymbol{\phi} = \{\phi_\alpha | \alpha \in \mathcal{A}\}$ denotes the set of potential functions defined on the cliques, and where $\boldsymbol{\theta} = \{\theta_\alpha | \alpha \in \mathcal{A}\}$ denotes the set of parameters associated with these potential functions, and where $Z_p$ is a normalization constant (the *partition function*). Given a disjoint variable partitioning, $\mathcal{C}$, the true *cluster conditional* of each variable cluster $C_i$ given its Markov blanket $MB_i$ is:

$$p(\mathbf{X}_{C_i}|\mathbf{X}_{MB_i} = \mathbf{x}_{MB_i}) \propto \exp\Big\{\sum_{D_\alpha \subseteq C_i} \theta_\alpha \phi_\alpha(\mathbf{X}_{D_\alpha}) + \sum_{D_\beta \subseteq B_i} \theta_\beta \phi_\beta(\mathbf{X}_{D_\beta \cap C_i}, \mathbf{x}_{D_\beta \cap MB_i})\Big\}, \tag{2}$$

where $B_i$ denotes the set of cliques that intersect with but are not contained in cluster $C_i$, and where the lower-case character $\mathbf{x}$ denotes a specific assignment to variable $\mathbf{X}$. Without loss of generality, we assume that all the potentials are positively weighted (i.e., $\theta > 0$) and the signs are subsumed in the potential functions.

Given a clique $D_\beta$, let $I_\beta$ denote the set of indices of clusters that have non-empty intersection with $D_\beta$. Let $I_{\beta i}$ denote $I_\beta \setminus i$. Finally, let us refer to the expectation of the potential $\phi_\beta(\mathbf{X}_{D_\beta})$ under the mean field cluster marginals indexed by $I_{\beta i}$ as a *peripheral marginal potential* of cluster $C_i$:

$$\phi'_\beta(\mathbf{X}_{D_\beta \cap C_i}, q_{I_{\beta i}}) \triangleq \langle \phi_\beta(\mathbf{X}_{D_\beta}) \rangle_{q_{I_{\beta i}}} = \int \phi_\beta(\mathbf{X}_{D_\beta \cap C_i}, \mathbf{x}_{D_\beta \cap MB_i}) q_{I_{\beta i}}(\mathbf{x}_{D_\beta \cap MB_i}) d\mathbf{x}_{D_\beta \cap MB_i}, \tag{3}$$

where $q_{I_{\beta i}}(\cdot) = \prod_{j \in I_{\beta i}} q_j(\mathbf{X}_{C_j})$.

Given the peripheral marginal potentials of all the cliques intersecting with cluster $C_i$, the GMF approximation to the cluster marginal of this cluster is:

$$q_i(\mathbf{X}_{C_i}) \propto \exp\Big\{\sum_{D_\alpha \subseteq C_i} \theta_\alpha \phi_\alpha(\mathbf{X}_{D_\alpha}) + \sum_{D_\beta \subseteq B_i} \theta_\beta \phi'_\beta(\mathbf{X}_{D_\beta \cap C_i}, q_{I_{\beta i}})\Big\}, \tag{4}$$

from which the isomorphism of the GMF approximation of the cluster marginal to the true cluster conditional (i.e., Eq. (2)) is apparent.

The definition of $\phi'_\beta$ makes it clear that the right-hand side of the mean field equation of cluster marginal $q_i$ depends only on a set of cluster marginals that are dependent on the Markov blanket variables of cluster $C_i$, which does not include $q_i$. Thus, Eqs. (3) and (4) constitute an asynchronous iteration procedure that loops over each cluster, calculating its peripheral marginal potentials using the current cluster marginals of its Markov blanket clusters, and then updating its own cluster marginal, until a fixed point is reached. This approach makes it straightforward to obtain update equations—all that is needed is to decide on a subgraph and a variable clustering, to identify the Markov blanket of each cluster, and to plug in the peripheral marginal potentials according to Eq. (4).

The definition of peripheral marginal potential is more general than the *mean field messages* defined in Xing et al. (2003), which can be viewed as a special case that applies to *cluster-factorizable potentials.* For other non-factorizable potentials, such as tabular potentials, peripheral marginal potentials are still well defined.

## 3 Bounds on GMF approximation

The quality of the GMF approximation depends critically on the choice of variable clustering of the graphical model. In the following we present a theorem that formally characterizes this relationship.

**Theorem 1 (GMF bound on KL divergence)**: *The Kullback-Leibler divergence between the GMF approximation to the joint posterior and the true joint posterior is bounded by the sum of the weights of potential functions associated with the cross-border cliques, up to some constants intrinsic to the graphical model:*

$$aW \leq KL(q\|p) \leq bW, \tag{5}$$

*where* $W = \sum_{D_\beta \subseteq \cup B_i} \theta_\beta$*, a and b are constants determined by the potential functions of the cross-border cliques (but independent to the potentials internal to the clusters.)*

A proof of this theorem is provided in the Appendix. The theorem of GMF bound provides a clear guideline for choosing a desirable partitioning of a general graphical model: empirically, it is desirable to break cliques associated with small weights while clustering variables; more systematically, we can use a graph partitioning algorithm to seek an optimal decomposition of the graph underlying the model. In the following, we explore several graph partitioning strategies on random graphs with pairwise potentials (each clique contains only two variables) to confirm and exploit Theorem 1 experimentally.

## 4 Variable clustering via graph partitioning

A wide variety of graph partitioning algorithms have been explored in recent years in a number of fields (e.g., Goemans and Williamson, 1995, Rendl and Wolkowicz, 1995). Given our focus on disjoint clusters in the GMF approach, these algorithms have immediate relevance to the problem of choosing clusters for inference. In this section, we describe the methods that we have explored.

### 4.1 Graph partition

Let $G(V, E, A)$ be a weighted undirected graph with node set $V = \{1, \ldots, n\}$, edge set $E$ and nonnegative weights $a_{ij}$, for $(i, j) \in E$ ($a_{ij} = 0$ if there is no edge between node $i$ and $j$; also $a_{ii} = 0, \forall i$). We refer to the symmetric matrix $A = \{a_{ij}\}$ as the *affinity matrix.* We equip the space of $n \times n$ matrices with the trace inner product $A \bullet B = \text{tr } AB$; let $A \succeq 0$ denote positive semidefiniteness ($A \succeq B$ denotes $A - B \succeq 0$); and let $A \geq 0$ denote elementwise non-negativity of $A$. The linear operator $\text{Diag}(a)$ forms a diagonal matrix from the vector $a$, and its adjoint operator $\text{diag}(A)$ yields a vector containing the diagonal elements of $A$. We denote by $e_k$ the vector containing $k$ ones.

#### 4.1.1 Equi-MinCut

We first consider graph partition (GP) problems based on minimum cuts. Given a graph $G(V, E, A)$ as described above, a classical formulation (Rendl and Wolkowicz, 1995) asks to partition the node set into $k$ disjoint subsets, $(S_1, \ldots, S_k)$, of specified sizes $m_1 \geq m_2 \geq \ldots \geq m_k$, $\sum_{j=1}^k m_j = n$, so as to minimize the total weight of the edges connecting nodes in distinct subsets of the partition. This is known as the minimum *k-cut* of $G$. In this paper, we concern ourselves with the special case of this problem in which all subsets have equal cardinality $m$, a problem that we refer to as *k equi-MinCut* (*k-MinC*). [1] Equi-MinCut avoids potentially skewed cuts on highly imbalanced graphs, and leads to a balanced distribution of computational complexity among clusters.

A $k$-way node partition can be represented by an *indicator matrix* $X \in \mathbb{R}^{n \times k}$ with the $j$-th column, $x_j = (x_{1j} \;\; x_{2j} \ldots x_{nj})^t$, being the *indicator vector* for the set $S_j$, $\forall j$:

$$x_{ij} = \begin{cases} 1 & : \quad \text{if } \; i \in S_j \\ 0 & : \quad \text{if } \; i \notin S_j \end{cases} .$$

Thus, *k-equipartitions* of a graph are in one-to-one correspondence with the set

$$\mathcal{F}_k = \{X : Xe_k = e_n, \; X^t e_n = m, \; x_{ij} = \{0, 1\}\}.$$

For each partition $X$, the total weight of the edges connecting nodes within cluster $S_i$ to its complement $\bar{S}_i$ is equal to $\frac{1}{2} x_i^t (D - A) x_i$, where $D = \text{Diag}(Ae_n)$. As a result, the total weight of the *k-cut* is

$$C_k = \sum_i \frac{1}{2} x_i^t (D - A) x_i = \frac{1}{2} \text{tr} X^t L X, \tag{6}$$

where $L \triangleq D - A$ is the *Laplacian matrix* associated with $G$.

[1] In combinatorial optimization, this problem is traditionally referred to as the *k-partition problem.* It is NP-hard, and to be distinguished from *unconstrained* minimum cut, which is *not* NP-hard.

Thus, $k$ equi-MinCut can be modeled as the following integer programming problem

$$(k\text{-}MinC) \quad MinC_k^* := \min\{\text{tr } X^t L X \ : \ X \in \mathcal{F}_k\}.$$

#### 4.1.2 Equi-MaxCut

We may also wish to find a $k$-partition that *maximizes* the total weight of the cut. This problem is known as the Max $k$-Cut in combinatorial optimization. Even without any size constraint this problem is NP-hard. In this paper, we again concern ourselves with a constrained version of the problem, in which all subsets have equal cardinality $m$. Thus we have the following *k equi-MaxCut* ($k$-$MaxC$) problem

$$\begin{aligned}(k\text{-}MaxC) \quad MaxC_k^* \ &:= \ \max\{\text{tr } X^t L X \ : \ X \in \mathcal{F}_k\} \\ &= \sum_i d_{ii} - \min\{\text{tr } X^t A X \ : \ X \in \mathcal{F}_k\}.\end{aligned}$$

We see that both $k$-$MaxC$ and $k$-$MinC$ are quadratic programs, and the relaxations that we consider will treat them identically.

#### 4.1.3 Weight matrices

The design of the affinity matrix has a fundamental impact on the results that are produced by graph partition algorithms. The naive choice in our case is to simply let $a_{ij} = 1$ when node $i$ and $j$ are connected in a graphical model, and let $a_{ij} = 0$ otherwise. Intuitively, an equi-MinCut using such an affinity matrix will capture more of the local dependency structure in the model, while an equi-MaxCut will lead to lower computational cost for inference in each cluster.

One can also partition the graphical model based on *coupling strength*, i.e., letting $a_{ij} = \theta_{ij}$, the weights of the pairwise potentials, so that an equi-MinCut results in clusters with strong intra-cluster coupling, whereas an equi-MaxCut produces a clustering with only weak couplings left in each cluster.

It also seems sensible to consider weighting schemes that favor large cuts with small coupling strength, or small cuts and large coupling strengths. We explore such a scheme by choosing weights that are inversely related to coupling strength.

The following table summarizes the various partition strategies explored in this paper, and the corresponding design of the affinity matrix.

Table 1: Graph partition schemes

| GP scheme | $k$-$MinC_a$ | $k$-$MinC_b$ | $k$-$MinC_c$ | $k$-$MaxC_a$ | $k$-$MaxC_b$ | $k$-$MaxC_c$ |
|---|---|---|---|---|---|---|
| $a_{ij}$ value | $\{1, 0\}$ | $\{\theta_{ij}, 0\}$ | $\{\frac{1}{\theta_{ij}}, 0\}$ | $\{1, 0\}$ | $\{\theta_{ij}, 0\}$ | $\{\frac{1}{\theta_{ij}}, 0\}$ |

### 4.2 Semi-definite relaxation of GP

Both *k equi-MinCut* and *k equi-MaxCut* are NP-hard. But there exist a variety of heuristics for finding approximate solutions to these problems (Frieze and Jerrum, 1995, Karisch and Rendl, 1998). In the sequel, we describe an algorithm that finds an approximate solution to $k$-$MinC$ and $k$-$MaxC$ using a semidefinite programming (SDP) relaxation (Karisch and Rendl, 1998).

#### 4.2.1 Semidefinite programming

Semidefinite programming (SDP) refers to the problem of optimizing a convex function over the convex cone of symmetric and positive semidefinite matrices, subject to linear equality constraints (Vandenberghe and Boyd, 1996). A canonical (primal) SDP takes the form:

$$(\textbf{SDP}) \quad \left\{ \begin{array}{lll} \underline{\min} & C \bullet X & \\ \underline{\text{s.t.}} & A_i \bullet X = b_i & \text{for } i = 1, \ldots, m \\ & X \succeq 0 & \end{array} \right.$$

Because of the convexity of the objective function and the feasible space, SDP problems have a single global optimum. With the development of efficient, general-purpose SDP solvers based on interior-point methods (e.g., SeDuMi (Sturm, 1999)), SDP has become a powerful tool in solving difficult combinatorial optimization problems. Here, we describe a simple SDP relaxation for solving graph partitioning problems.

#### 4.2.2 SDP relaxation of GP

We now derive a semidefinite relaxation for GP. For simplicity, we illustrate only for $k$-$MinC$; $k$-$MaxC$ follows similarly with the appropriate change to the objective.

The first step in SDP relaxation involves replacing $X^t L X$ with tr $LY$, where $Y$ is equal to $XX^t$; this *linearizes* the objective. Let us define the set $\mathcal{T}_k$:

$$\mathcal{T}_k := \{Y : \ \exists X \in \mathcal{F}_k \text{ such that } \ Y = XX^t\}.$$

Thus $k$-$MinC$ reads: $MinC_k^* := \min\{\text{tr } LY \ : \ Y \in conv(\mathcal{T}_k)\}$.

Note that due to linearization of the objective, our feasible set can be rewritten as the convex hull of the original set $\mathcal{T}_k$. The next step is to approximate the convex hull of $\mathcal{T}_k$ by outer approximations that can be handled efficiently. Karisch and Rendl (1998) describe a nested sequence of outer approximations for GP that leads from the well-known eigenvalue bound of Donath and Hoffman to increasingly accurate bounds. Omitting details, one of their relaxation schemes results in the following SDP relaxation for $k$-$MinC$:

Table 2: GP performance. (default: $K$-means rounding; rp: random projection rounding)

| | Equi-MinCut | | | Equi-MaxCut | | | | |
|---|---|---|---|---|---|---|---|---|
| $k$ | lower-b | feas. $X$ | f/b | upper-b | feas. $X$ | f/b | feas. $X$ (rp) | f/b (rp) |
| 3 | 34 | 38 | 1.10±0.03 | 78 | 75 | 0.96±0.02 | 71 | 0.91±0.04 |
| 4 | 41 | 45 | 1.09±0.02 | 82 | 80 | 0.97±0.02 | 74 | 0.90±0.04 |
| 6 | 52 | 55 | 1.06±0.03 | 83 | 81 | 0.97±0.01 | 77 | 0.93±0.02 |
| 8 | 59 | 61 | 1.03±0.02 | 83 | 82 | 0.99±0.01 | 79 | 0.95±0.02 |
| 3 | 73 | 77 | 1.05±0.02 | 122 | 119 | 0.97±0.01 | 113 | 0.92±0.03 |
| 4 | 86 | 90 | 1.05±0.02 | 135 | 130 | 0.97±0.01 | 122 | 0.91±0.03 |
| 6 | 104 | 207 | 1.03±0.01 | 140 | 137 | 0.98±0.01 | 128 | 0.91±0.02 |
| 8 | 116 | 118 | 1.02±0.01 | 140 | 138 | 0.99±0.01 | 131 | 0.93±0.01 |

$$(P) \quad \begin{cases} \underline{\max} & \frac{1}{2}\text{tr } LY \\ \underline{\text{s.t.}} & \text{diag}(Y) = e_n \\ & Ye_n = me_n \\ & Y \geq 0 \quad \text{elementwise} \\ & Y \succeq 0, \; Y = Y^t \end{cases}$$

(P) is an SDP and can be solved by an interior-point solver such as SeDuMi.

#### 4.2.3 Finding a closest feasible solution

While in some cases a bound is the major goal of a relaxation, in our case we require that the relaxation give us a (feasible) solution. In particular, the optimal solution of problem $(P)$ is in general not feasible for the original GP problem, and we need to recover from the approximate solution a closest feasible solution, $X$, to the original GP problem. We use the following scheme in this paper.

- From the relaxed solution $Y$, find a decomposition $Y = X'X'^t$ via SVD (note that $X'$ is usually full rank rather than of rank $k$ as in the feasible case).
- Treat each row in $X'$ as a point in $\mathbb{R}^n$; cluster these points using a variant of the standard $K$-means algorithm that finds equi-size clusters. (We use multiple restarts and pick the result with the best cut value).
- Complete the feasible index matrix $X$: $x_{ij} = 1$ iff row $i$ of $X'$ gets assigned to cluster $j$.

This rounding scheme is related to the randomized projection heuristic studied by Goemans and Williamson (1995) in their work on Max-Cut. In this approach, the label (-1 or +1) of each vector is chosen according to whether the vector is above or below a randomly chosen hyperplane passing through the origin. Frieze and Jerrum (1995) generalized this scheme to max $k$-cut. Rendl and Wolkowicz (1995) proposed another alternative involving a first-order Taylor expansion of the cost function around the relaxed $X'$. However, these schemes make it difficult to enforce size constraints on the clusters, and occasionally produce artifacts such as having an empty cluster. Empirically, we have found that a $K$-means heuristic usually leads to superior and often near-optimal results.

## 5 Experiments and Results

In this section, we combine graph partitioning with the GMF algorithm to perform inference on randomly generated undirected graphical models with singleton and pairwise potentials. We analyze three aspects of the overall procedure—the quality of graph partition, the accuracy of approximate marginal probabilities, and the tightness of lower bounds on the log partition function.

For each trial, we use a random graph of 24 nodes [2] and specify the distribution $p(\mathbf{x}|\theta)$ by making a random choice of the model parameter $\theta$ from a uniform distribution $\mathcal{U}(a, b)$. For single node weights $\theta_i$, we set $a = -w_{\text{obs}}$ and $b = -w_{\text{obs}}$; for pairwise weights $\theta_{ij}$, we set $a = -w_{\text{coup}}$, $b = 0$ for *repulsive* coupling, $a = -w_{\text{coup}}$, $b = w_{\text{coup}}$ for *mixed* coupling, and $a = 0$, $b = w_{\text{coup}}$ for *attractive* coupling, respectively.

### 5.1 Partitioning random graphs

Our graphs are generated by sampling an edge with probability $p$ for each pair of nodes. Table 2 summarizes the performance (over 100 trials) of various graph partition schemes on random graphs. To assess performance, we compute the ratio $f/b$ between the feasible cut and the bound of the cut provided by the SDP relaxation (the optimal solution must fall between $f$ and $b$). In the top panel, we show results for partitioning unweighted graphs with $p = 0.3$ into $k = 3, 4, 6$, and 8 clusters. The bottom panel shows results for partitioning denser unweighted graphs with $p = 0.5$. Partitions on weighted graphs show similar performance.

We see that the SDP-based GP provides very good and stable partitioning results, usually no worse than 10% of the optimal cut values, and often within 5%. Note also that the $K$-means rounding scheme outperforms the random projection rounding (rp).

[2] Indeed, a standard SDP solver can readily handle larger graphs (e.g., with more than 100 nodes). But the exact solutions of the singleton marginals of larger graphs are very expensive to compute, which makes it difficult to obtain good estimates of the inference error.

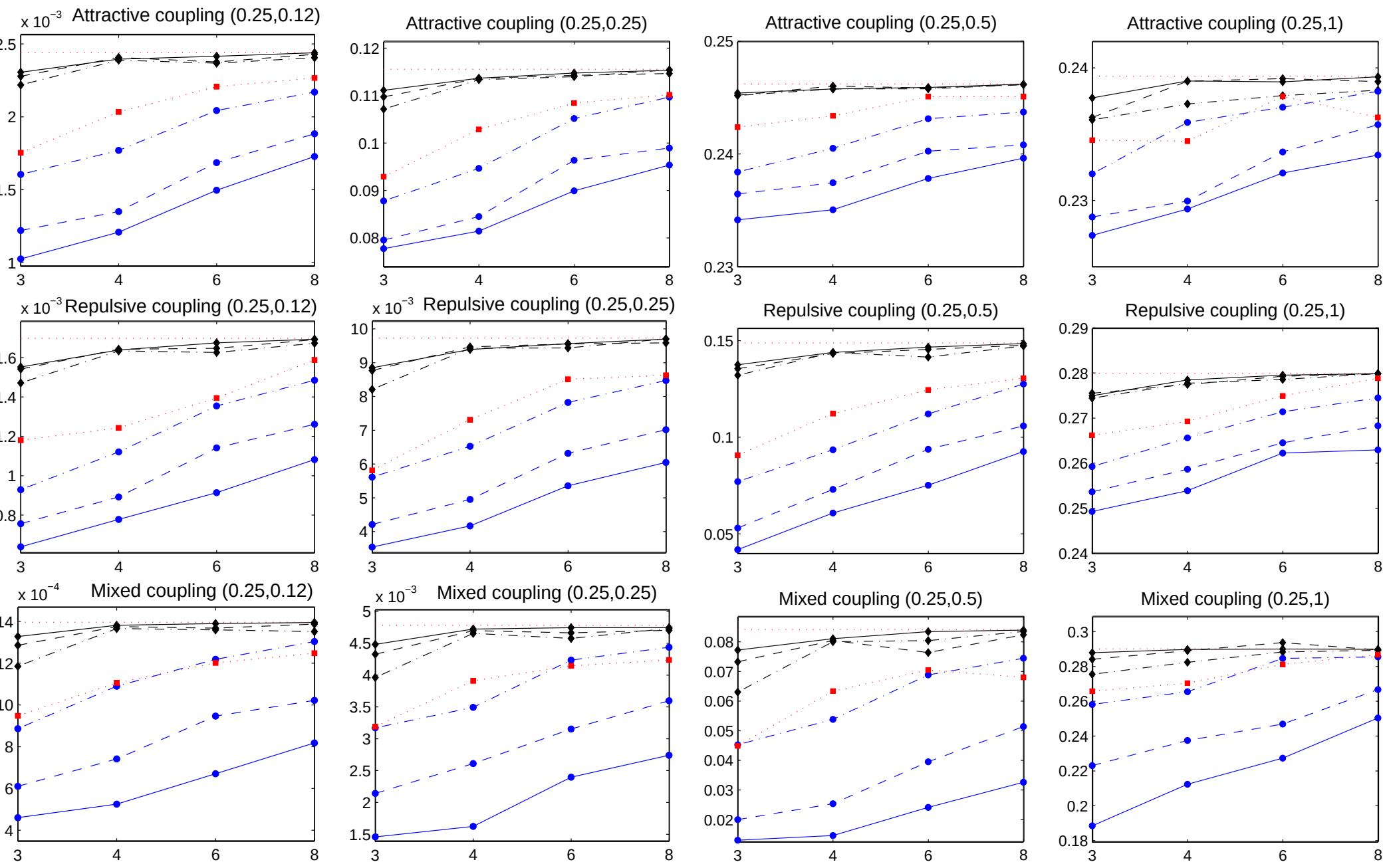

Figure 1: $\ell_1$ errors of singleton marginals on random graphs, with different coupling types, and strengths. Each experiment is based on 20 trials. The sampling ranges of the model parameters for each set of trials are specified on top of each graph as $(w_{\text{obs}}, w_{\text{coup}})$. ($x$-axis: the number of clusters; $y$-axis: the $\ell_1$ error; solid lines: cut based on $\theta_{ij}$-weighted $A$; dashed lines: cut based on unweighted $A$; dashed-dot lines: cut based on $1/\theta_{ij}$-weighted $A$; lines with diamond symbols: equi-MaxCut (black); lines with round-dot symbols: equi-MinCut (blue); dotted line with square symbols: random cut (red). For reference, the dotted (red) line with no symbol marks the baseline error of naive mean field.)

### 5.2 Single-node marginals

We compared performance of GMF using different graph partition schemes with regard to the accuracy on single-node marginals. We used all six GP strategies summarized in Table 1, as well as a random clustering scheme. To assess the error, we use an $\ell_1$-based measure as described in Xing et al. (2003). The exact marginals are obtained via exhaustive enumeration. We used graphs of two different densities in our experiments: *moderately connected* graphs, with treewidth $12 \pm 1$, more than an order of magnitude greater than the largest cluster to be formed; and *densely connected* graphs, with treewidth $16 \pm 1$. For reasons of space, we show only results for the moderately connected graphs.

Figure 1 shows that for all variable clusterings, GMF almost always improves over the naive mean field. As expected, equi-MinCut always provides better results than other partition strategies. In particular, equi-MinCut based on coupling strength yields the best results (consistent with Theorem 1), followed by equi-MinCut based on node degree, then equi-MinCut that cuts the least number of heavy edges. This suggests that, to better approximate the true marginals, it is important to capture strong couplings within clusters. Equi-MaxCut fares less well; indeed, it is worse than a random cut in most cases. It is worth noting, however, that cutting lightweight edges (i.e., maximizing the sum of $\frac{1}{\theta_{ij}}$ across clusters) leads to better performance than degree- or coupling-based cuts.

For denser graphs (results not shown), the performance gap between different clustering schemes becomes smaller, but the trend and the relative order remain the same.

### 5.3 Bounds on log partition function

Figure 2 shows the lower bounds of the log partition functions given by the GMF approximations. Comparing to Figure 1, we see that there is a good correspondence between the performance on approximating marginals and the tightness of the lower bound, a reassuring result in the context of mean-field algorithms.

## 6 Discussion

We have investigated combinations of graph partition algorithms with the generalized mean field algorithm, which allows mean field approximations to be optimized over both parameter space and variable partition space in an autonomous fashion. We proved that the quality of the GMF approximation is bounded by the total absolute weight of the potentials of the disrupted cliques due to the disjoint variable clustering underlying GMF approximation. Empirically, we confirmed that although all graphs partitions lead to improvement over a naive mean field approximation, a minimal cut equipartition clearly yields the best GMF approximation, measured both by singleton marginals and lower bounds of the true log partition function. Moreover, there is a good association between the qualities of the approximate marginals and lower bounds.

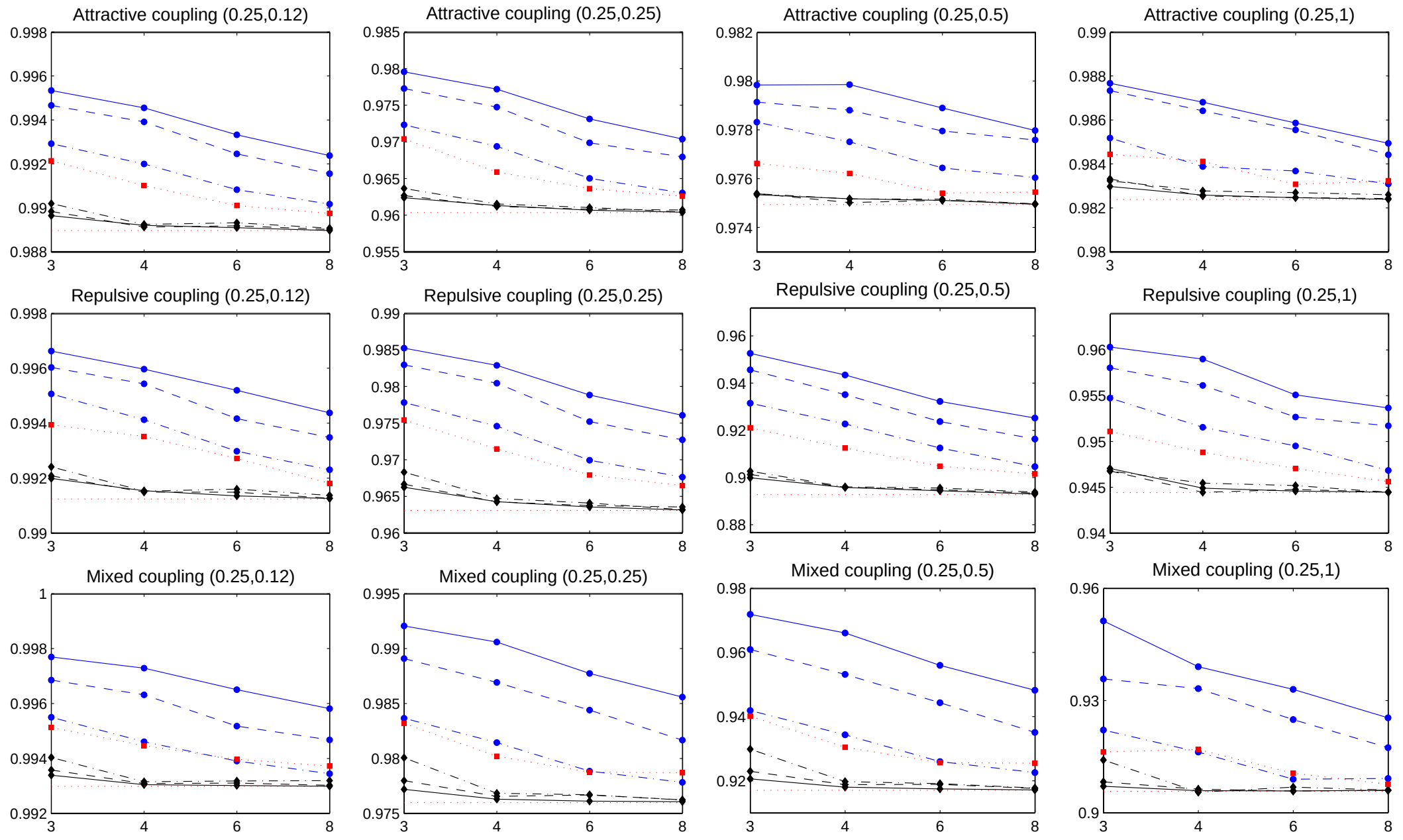

Figure 2: Accuracy of the lower bound on the log partition function. The ordering of the panels and the legends are the same as in Fig 1, except that the $y$-axis now corresponds to the ratio of the lower bound of the log partition function due to GMF versus the true log partition function.

This paper represents an initial foray into the problem of choosing clusters for cluster-based variational methods. There is clearly much more to do. First, we should note that we are far from the ideal approach, where we base the clustering criterion on the ultimate goal—that of obtaining accurate estimates of marginal probabilities. This is of course an ambitious goal to aim for, and in the near term it seems advisable to attempt to find effective surrogates. In particular, we do not want the problem of choosing clusters to be as computationally complex as the inference problem that we wish to solve! (Fortunately, many efficient solvers are available to solve the GP problem nearly optimally via SDP or spectral relaxation.) We should consider surrogates that involve more general combinations of parameter values along cuts. In particular, we found little support for the use of maximum cuts in our experiments, but perhaps if we search for large cuts along which the parameter values are uniformly small we will have more success in this regard. In general, we might ask for a surrogate that aims to capture both the setting under which mean field approximations are effective, and the setting under which important local dependencies can be treated tractably.

Note also that we have focused on partitioning methods that decompose a large graphical model into clusters of equal size. With no prior knowledge of the local connectivity within the clusters, this equal-size heuristic seems reasonable; we wish to roughly distribute resources equally to each cluster (e.g., to balance the load in a parallel computing setting). Again, however, it would be useful to explore surrogates that attempt to capture local connectivity in the clustering criterion.

In light of Theorem 1, there are a number of extensions of the research reported here that potentially lead to further improved GMF approximation. First, Eq. 8 in the appendix suggests that it may be advantageous to use other weighting schemes, such as the entropy-like clique weights (expected potentials) $\langle \phi_\beta \rangle_q$, and seek a partition that minimizes the sum of expected cross-border potentials. Obviously, exact computation of the entropy-like weights requires the true joint distribution, and is thus infeasible. We may approximate the expected potential of each clique by replacing the true marginal distribution of the variables in the corresponding clique with a naive mean-field-like approximation to the marginal: $q(X_{D_\beta}) \propto \exp\{\theta_\beta \phi_\beta(X_{D_\beta})\}$; this turns the computation of the expectation into a local computation. It is possible to use a algorithm that iterates between GMF (to update the marginal $q(\cdot)$) and GP (to update the partition).

Another possible extension is to replace the disjoint clustering with the *tree-connected clustering*. The term $W$ in the GMF bound can be also viewed as the total weight of the disrupted cliques (with respect to the original graph) in the subgraph underlying a GMF approximation. Thus, we may further reduce $W$ by departing from the completely disjoint clustering to tree-connected clusters, in which we connect all the disjoint clusters resulting from a graph partition using a tree whose nodes are clusters. The link between every pair of connected clusters is chosen to be the maximally weighted clique shared by the clusters. Such a tree can be easily obtained via constructing a maximal spanning tree of variable clusters. The motivation of using tree-connected clusters rather than arbitrary subgraphs to approximate the true joint distribution is that under such a subgraph, the message-passing-based GMF algorithm described earlier is still guaranteed to converge and yield a set of globally consistent approximate cluster marginals.

## A Proof of the GMF bound theorem

**Proof.**

According to the GMF theorem, the GMF approximation to $p(\mathbf{X})$ is

$$\begin{aligned} q(\mathbf{X}) &= \prod_i q(\mathbf{X}_{C_i}) \\ &= \frac{1}{Z_q}\exp\Big\{\sum_i\sum_{D_\alpha\subseteq C_i}\theta_\alpha\phi_\alpha(\mathbf{X}_{D_\alpha}) + \sum_i\sum_{D_\beta\subseteq B_i}\theta_\beta\phi'_\beta(\mathbf{X}_{D_\beta\cap C_i})\Big\} \\ &= \frac{1}{Z_q}\exp\Big\{\sum_\alpha\theta_\alpha\phi_\alpha(\mathbf{X}_{D_\alpha}) \\ &\qquad - \sum_{D_\beta\subseteq\cup B_i}\theta_\beta\phi_\beta(\mathbf{X}_{D_\beta}) + \sum_{D_\beta\subseteq\cup B_i}k_\beta\theta_\beta\phi'_\beta(\mathbf{X}_{D_\beta\cap C_i})\Big\} \\ &= \frac{1}{Z_q}\exp\Big\{\sum_\alpha\theta_\alpha\phi_\alpha(\mathbf{X}_{D_\alpha}) + \sum_{D_\beta\subseteq\cup B_i}\theta_\beta\big(k_\beta\phi'_\beta(\mathbf{X}_{D_\beta\cap C_i}) - \phi_\beta(\mathbf{X}_{D_\beta})\big)\Big\}, \end{aligned} \tag{7}$$

where $k_\beta = |I_\beta|$ is the number of clusters intersecting with clique $\beta$ (note that for simplicity, we omit the argument $q_{I_{\beta i}}$ in the peripheral marginal potentials). Thus, the KL divergence between $q$ and $p$ is:

$$\begin{aligned} \mathrm{KL}(q\|p) &= \int_{\mathbf{x}} q(\mathbf{x})\log\frac{q(\mathbf{x})}{p(\mathbf{x})}d\mathbf{x} \\ &= \sum_{D_\beta\subseteq\cup B_i}\theta_\beta\Big(k_\beta\big\langle\phi'_\beta(\mathbf{X}_{D_\beta\cap C_i})\big\rangle_q - \big\langle\phi_\beta(\mathbf{X}_{D_\beta})\big\rangle_q\Big) - \log\frac{Z_q}{Z_p} \\ &= \sum_{D_\beta\subseteq\cup B_i}\theta_\beta(k_\beta-1)\big\langle\phi_\beta(\mathbf{X}_{D_\beta})\big\rangle_q - \log Z_q + \log Z_p. \end{aligned} \tag{8}$$

Now, let $\phi_{\beta,\max} = \max_{\mathbf{x}}\phi_\beta(\mathbf{x}_{D_\beta})$, $\phi_{\beta,\min} = \min_{\mathbf{x}}\phi_\beta(\mathbf{x}_{D_\beta})$, we have, $\phi_{\beta,\min} \le \langle\phi_\beta(\mathbf{X}_{D_\beta})\rangle_q \le \phi_{\beta,\max}$. Define $a_\phi = \min_{D_\beta\subseteq\cup B_i}(k_\beta-1)\phi_{\beta,\min}$, and $b_\phi = \max_{D_\beta\subseteq\cup B_i}(k_\beta-1)\phi_{\beta,\max}$, then (since all the $\theta$s are positive by definition),

$$a_\phi\sum_{D_\beta\subseteq\cup B_i}\theta_\beta \le \sum_{D_\beta\subseteq\cup B_i}\theta_\beta(k_\beta-1)\big\langle\phi_\beta(\mathbf{X}_{D_\beta})\big\rangle_q \le b_\phi\sum_{D_\beta\subseteq\cup B_i}\theta_\beta. \tag{9}$$

To bound the log partition function, we find that

$$
\begin{aligned}
Z_q &= \sum_{\mathbf{x}} \exp\Big\{\sum_{\alpha} \theta_\alpha \phi_\alpha(\mathbf{x}_{D_\alpha})\Big\} \times \\
&\quad \exp\Big\{\sum_{D_\beta \subseteq \cup B_i} \theta_\beta \big(k_\beta \phi'_\beta(\mathbf{x}_{D_\beta \cap C_i}) - \phi_\beta(\mathbf{x}_{D_\beta})\big)\Big\} \\
&\leq \sum_{\mathbf{x}} \exp\Big\{\sum_{\alpha} \theta_\alpha \phi_\alpha(\mathbf{x}_{D_\alpha})\Big\} \times \exp\Big\{b_Z \sum_{D_\beta \subseteq \cup B_i} \theta_\beta\Big\} \\
&= Z_p \exp\Big\{b_Z \sum_{D_\beta \subseteq \cup B_i} \theta_\beta\Big\}, \qquad (10)
\end{aligned}
$$

where

$$
\begin{aligned}
b_Z &= \max_\beta (k_\beta \phi_{\beta,\max} - \phi_{\beta,\min}) \\
&= \max_\beta \big((k_\beta - 1)\phi_{\beta,\max} + (\phi_{\beta,\max} - \phi_{\beta,\min})\big). \qquad (11)
\end{aligned}
$$

Similarly,

$$
Z_q \geq Z_p \exp\Big\{a_Z \sum_{D_\beta \subseteq \cup B_i} \theta_\beta\Big\}, \qquad (12)
$$

where

$$
a_Z = \min_\beta \big((k_\beta - 1)\phi_{\beta,\min} + (\phi_{\beta,\min} - \phi_{\beta,\max})\big). \qquad (13)
$$

Thus,

$$
\log Z_p + a_Z \sum_{D_\beta \subseteq \cup B_i} \theta_\beta \leq \log Z_q \leq \log Z_p + b_Z \sum_{D_\beta \subseteq \cup B_i} \theta_\beta. \qquad (14)
$$

Putting these together, we have

$$
aW \leq \mathrm{KL}(q\|p) \leq bW, \qquad (15)
$$

where $a = \max(0,\ a_\phi - b_Z)$ and $b = b_\phi - a_Z$.

In the special case where $k_\beta = k$, for all $\beta$ (e.g., all pairwise potentials), $a_\phi - b_Z = (k-1)\min_{\beta,\beta'}(\phi_{\beta,\min} - \phi_{\beta',\max}) + \min_\beta(\phi_{\beta,\min} - \phi_{\beta,\max}) \geq k \min_{\beta,\beta'}(\phi_{\beta,\min} - \phi_{\beta',\max})$, and $b = b_\phi - a_Z \leq k \max_{\beta,\beta'}(\phi_{\beta,\max} - \phi_{\beta',\min}) \equiv k\Delta_\phi$. Since KL divergence is always nonnegative, we have

$$
0 \leq \mathrm{KL}(q\|p) \leq k\Delta_\phi W. \qquad (16)
$$

■